\documentclass{article}

\usepackage[preprint]{neurips_2024}

\usepackage[utf8]{inputenc} 
\usepackage[T1]{fontenc}    
\usepackage{hyperref}       
\usepackage{url}            
\usepackage{booktabs}       
\usepackage{amsfonts}       
\usepackage{nicefrac}       
\usepackage{microtype}      
\usepackage{xcolor}         
\usepackage{algorithm}
\usepackage{algorithmic}
\usepackage{amsmath}
\usepackage{graphicx}
\usepackage{subcaption} 

\title{Improving Group Fairness in Knowledge Distillation via Laplace Approximation of Early Exits}

\author{
  Edvin Fasth \\
  EESC \\
  KTH Royal Institute of Technology\\
  \texttt{edvinfa@kth.se} \\
   \And
  Sagar Singh\\
  CSE \\
  IIT Bombay\\
  \texttt{sagarsingh@cse.iitb.ac.in}
}

\begin{document}

\maketitle

\begin{abstract}
Knowledge distillation (KD) has become a powerful tool for training compact student models using larger, pretrained teacher models, often requiring less data and computational resources. Teacher models typically possess more layers and thus exhibit richer feature representations compared to their student counterparts. Furthermore, student models tend to learn simpler, surface-level features in their early layers. This discrepancy can increase errors in groups where labels spuriously correlate with specific input attributes, leading to a decline in group fairness even when overall accuracy remains comparable to the teacher.
To mitigate these challenges, Early-Exit Neural Networks (EENNs), which enable predictions at multiple intermediate layers, have been employed. Confidence margins derived from these early exits have been utilized to reweight both cross-entropy and distillation losses on a per-instance basis.

In this paper, we propose that leveraging Laplace approximation-based methods to obtain well-calibrated uncertainty estimates can also effectively reweight challenging instances and improve group fairness. We hypothesize that Laplace approximation offers a more robust identification of difficult or ambiguous instances compared to margin-based approaches. To validate our claims, we benchmark our approach using a Bert-based model on the MultiNLI dataset.
\end{abstract}
  
\section{Introduction}

Knowledge Distillation (KD) \cite{hinton2015distilling} offers a sample-efficient approach for training a student model by leveraging a pretrained teacher model. The training loss for the student is a combination of the standard cross-entropy loss calculated on the ground-truth labels and a knowledge distillation (KD) loss that encourages the student's predictions to align with the "soft targets" provided by the teacher. 
Let $D={(x_i, y_i)}_{i=1}^{N} $ represent the training dataset, 
$\mathcal{L}_{student} = (1 - \alpha) \mathcal{L}_{CE}(f_s(x_i), y_i) + \alpha \mathcal{L}_{KD}(f_s(x_i; T), f_t(x_i; T))$,
with T being the temperature parameter usually kept high during training. 
This process effectively transfers the teacher's generalization capabilities to the student. A significant advantage of KD is the flexibility to choose a student model with significantly fewer parameters than the often computationally intensive teacher, thereby facilitating easier deployment.

One issue identified in the original knowledge distillation paper is the occurrence of bias during the training process. Since the student model learns from the teacher’s output probabilities, any biases present in the teacher’s predictions can be inherited—or even amplified—by the student. This becomes particularly problematic in settings where the teacher model has overfit to spurious correlations in the data or exhibits imbalanced performance across subgroups. As a result, the distillation process may unintentionally reinforce unfair or misleading patterns, highlighting the need for strategies that account for such biases during student training. 

Reweighting losses for spuriously correlated groups has been shown to improve group fairness metrics. However, this approach typically requires prior knowledge of the group during training and lacks adaptivity based on the student model's current performance on challenging groups. To address these limitations, Tiwari et al. \cite{tiwari_using_2024} observed that student models often exhibit overconfidence on difficult instances in their early layers. They proposed utilizing Early-Exit Neural Networks (EENNs) to obtain uncertainty estimates from these intermediate layers and subsequently employing this information to dynamically reweight both the teacher's distillation loss and the student's cross-entropy loss for each training sample.

Meronen et al. \cite{meronen2024fixing} proposed an alternative method for obtaining well-calibrated uncertainty estimates from EENNs. Their approach treats the final layer of the classification heads in early exit blocks as Bayesian parameters, employing Laplace approximations. This offers a computationally efficient way to estimate the uncertainty associated with making a prediction at an early exit. Furthermore, this method tends to yield high uncertainty for difficult instances, attributed to potential model misspecification.

In this paper, we investigate whether this Laplace approximation-based uncertainty estimation in early exit layers can be effectively used to reweight challenging training instances.

Section \ref{sec:related} outlines the work of Tiwari et al. \cite{tiwari_using_2024} as the primary motivation for our approach. Section \ref{sec: Laplace} details the Laplace approximation method used for uncertainty estimation. The experimental setup and results are presented in Section \ref{sec: expt}. Finally, Section \ref{sec: conc} provides concluding remarks.

\subsection{Related Works \label{sec:related}}
\begin{figure}
    \centering
    \includegraphics[width=0.5\linewidth]{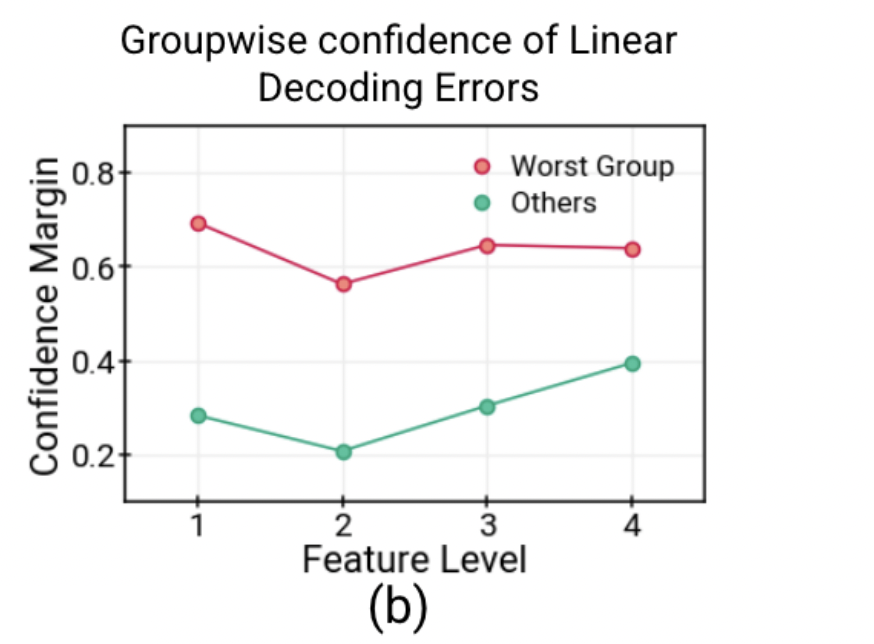}
    \caption{Confidence margin at different feature levels for the worst-performing group compared to other groups in the original DEDIER paper.}
    \label{fig:worst-group}
\end{figure}

\paragraph{Mediating Bias Using Early Readouts}
Our work closely builds upon the approach by Tiwari et al. to mitigate featural bias in student models \cite{tiwari_using_2024}.
To address the issue of bias propagation during knowledge distillation, Tiwari et al. introduced the DEDIER model. Their method adds the student model with an auxiliary network designed to estimate the confidence of early-layer predictions. This auxiliary network generates instance-level uncertainty estimates, which are then used to dynamically reweight the losses from training samples during distillation. They assign higher weights to instances with low  uncertainty predictions and thus overconfident instances in the early exits, then multiply these weights with KD loss from the teacher and reduces the impact of potentially biased or misleading data points. DEDIER thus introduces a more debiased training process that maintains the student model's efficiency and compactness while achieving improved performance on classification tasks. The overall loss was given by:
\[ \mathcal{L}_{student} =
          \sum_{D_w}^{} (1- \lambda) \cdot l_{ce}  \\
          + \lambda \cdot \textbf{wt} \cdot l_{ke} \] 
          where \(\bf{wt} = \exp^{\beta.\bf{cm}.\alpha} \) and \(\bf{cm(p)} = p_{max} - \max_{p_k \in \bf{p} - p_{max}} p_k \). The parameters \(\beta\) and \( \alpha\) are hyperparameters. 
While DEDIER offers a more robust framework compared to standard knowledge distillation, it remains reliant on the accuracy of the auxiliary network's confidence margins. This dependency has been identified as a potential limitation. Specifically, within subgroups of the dataset where classification performance is lowest, the prediction confidence margins tend to be higher than the average confidence across the dataset, as illustrated in Figure \ref{fig:worst-group}. In the next section we'll see how to use laplace approximation as another approach towards uncertainty estimation in early exits and how to use them for adjusting the loss function. 

\section{Laplace Approximation for uncertainty estimates \label{sec: Laplace}}

Identifying the need for better prediction of confidence margins, Meronen et al. \cite{meronen2024fixing} targeted overconfidence and miscalibration in their intermediate classifiers. To do so, the paper proposed a probabilistic framework that enhanced the uncertainty quantification at each exit point. By applying last-layer Laplace approximations and model-internal ensembling, the model captured both aleatoric and epistemic uncertainties more accurately. This led to improved calibration and more trustworthy confidence scores, which in turn allowed for better decision-making under computational budget constraints. The proposed method demonstrated superior performance on many benchmark datasets (CIFAR-100, ImageNet, Caltech-256) across multiple metrics, including accuracy, negative log predictive density (NLPD), and expected calibration error (ECE), without retraining or significant computational overhead.

The original DEDIER paper defines the confidence margin according to Eq. \ref{eq:cm}

\begin{equation}
p_{\text{max}} = \max_{p_k \in \mathbb{P}} p_k
\label{eq:pmax}
\end{equation}

\begin{equation}
\mathbf{cm}(\mathbf{p}) = p_{\text{max}} - \max_{p_k \in \mathbb{P} \setminus \{p_{\text{max}}\}} p_k
\label{eq:cm}
\end{equation}

This is used to re-weight the training samples for samples where the auxiliary network in the student model does not produce the correct label, according to Eq. \ref{eq:wt}

\begin{equation}
wt = 
\begin{cases}
1, & \text{if } y^{(\text{aux})} = y \\
e^{\beta \cdot \mathrm{cm}(\mathbf{p}^{(\text{aux})})^\alpha}, & \text{otherwise}
\end{cases}
\label{eq:wt}
\end{equation}

Instead of learning point estimates for the parameters in the auxiliary network, we apply a Laplace approximation over the final linear layer. Specifically, we approximate the posterior over the output weights of the auxiliary network using the Bayesian update rule from Eq.~\ref{eq:bayes_rule}:

\begin{equation}
\label{eq:bayes_rule}
p(\theta \mid \mathcal{D}_{\text{train}}) = 
\frac{p(\mathcal{D}_{\text{train}} \mid \theta)\, p(\theta)}
{\int p(\mathcal{D}_{\text{train}} \mid \theta)\, p(\theta) \, d\theta}
= \frac{\text{[likelihood]} \times \text{[prior]}}{\text{[model evidence]}}
\end{equation}

In practice, we apply this approximation to the final weight matrix $\hat{\mathbf{W}}_{\text{MAP}}$ of the auxiliary linear classifier. Let $\hat{\boldsymbol{\phi}}_i$ denote the BERT feature vector (CLS token) for input $\mathbf{x}_i$. The predictive distribution over the logits $\mathbf{z}_i$ is then:

\begin{equation}
\label{eq:conditional_gaussian}
p(\mathbf{z}_i \mid \mathbf{x}_i) = \mathcal{N}\left(\hat{\mathbf{W}}^\top_{\text{MAP}} \hat{\boldsymbol{\phi}}_i,\; \hat{\boldsymbol{\phi}}_i^\top \Sigma_{\phi} \hat{\boldsymbol{\phi}}_i \cdot \mathbf{I}_C \right)
\end{equation}

Here, $\Sigma_{\phi}$ is the empirical covariance of the feature vectors $\hat{\boldsymbol{\phi}}_i$, and $\mathbf{I}_C$ is the identity matrix over output classes. The resulting variance term reflects uncertainty from both the features and the weight posterior. The covariance $\Sigma_{\phi}$ is computed as:

\begin{equation}
\Sigma_{\phi} = \frac{1}{N-1} \sum_{i=1}^{N} (\hat{\boldsymbol{\phi}}_i - \bar{\boldsymbol{\phi}})(\hat{\boldsymbol{\phi}}_i - \bar{\boldsymbol{\phi}})^\top
\end{equation}

To ensure numerical stability and positive-definiteness, we add a small ridge term to $\Sigma_{\phi}$ as in the Laplace approximation. Samples from the predictive distribution are drawn using Monte Carlo, and the entropy of the resulting averaged softmax distribution is used to weight the loss function.

In order to make uncertainty predictions more accurate, we aim to integrate the work of Meronen et al. into the original DeDIER model.

\section{Experiments and results \label{sec: expt}}
\subsection{Datasets}
The MultiNLI dataset \cite{williams2018broad} was used for classification. MultiNLI is a benchmark dataset designed to evaluate a model’s ability to perform natural language inference (NLI), which involves determining whether a given premise sentence entails, contradicts, or is neutral with respect to a hypothesis sentence. The dataset encompasses examples from ten distinct genres of both written and spoken English, including fiction, government reports, telephone conversations, and popular culture articles, thus providing a diverse and challenging testbed for robust generalization. Each premise-hypothesis pair is manually annotated and labeled as "entailment," "neutral," or "contradiction." This dataset is widely adopted within the NLI research community and was also employed in the original DeDiER paper, which facilitates direct benchmarking and comparison of results.

\subsection{Model Architecture And Training}
The model is based on the original DeDIER approach, with some important modifications made to the confidence marings used in the auxiliary network.
\begin{algorithm}
\caption{The original DeDIER approach: learning student $S$ given dataset $\mathcal{D} = \{(x_i, y_i)\}_{i=1}^N$ and teacher $T$.}
\textbf{Hyperparameters:} Distillation loss fraction $\lambda$, early exit depth $d$, parameters $\alpha$ and $\beta$ for weight calculation.
\begin{algorithmic}[1]
\STATE $S = h(g_d(\cdot))$: split student into early layers $g_d$ of depth $d$ and deeper layers $h$
\STATE Let $h_{\text{aux}}(\cdot)$ be the auxiliary network
\STATE Augment dataset $\mathcal{D}$ with weights:
\[
\mathcal{D}_w \leftarrow \{(x_i, y_i, wt_i)\}_{i=1}^n, \quad \text{where } wt_i = 1
\]
\FOR{$e = 1$ to $E$}
    \STATE Train student $S$ using loss from Eq.~5
    \IF{$e \bmod L = 0$}
        \STATE Train $h_{\text{aux}}(g_d(\cdot))$ for $R$ epochs on $\mathcal{D}$ using cross-entropy
        \FOR{each $(x, y, wt) \in \mathcal{D}_w$}
            \STATE $y^{(\text{aux})} \leftarrow h_{\text{aux}}(g_d(x))$
            \STATE Update $wt$ according to Eq.~4
        \ENDFOR
    \ENDIF
\ENDFOR
\end{algorithmic}
\end{algorithm}

With Laplace implemented, this is our new version:

\begin{algorithm}
\caption{DeDiER with auxiliary projector and Laplace margins on MultiNLI}
\textbf{Inputs:} Teacher $T$, Student $S$, Projector $h_{\text{aux}}$, dataset $\mathcal{D}$, hyperparameters $\alpha$, $\beta$, number of epochs $E$.

\begin{algorithmic}[1]
\STATE Pre-train or load teacher model $T$ on $\mathcal{D}$
\STATE Initialize student $S$ and auxiliary projector $h_{\text{aux}}$
\FOR{$e = 1$ to $E$}
    \FOR{each minibatch $(x, y, m)$ from $\mathcal{D}$}
        \STATE Compute features $\phi = T_{\text{BERT}}(x)$ (CLS embedding)
        \STATE Train projector $h_{\text{aux}}$ using cross-entropy on $(\phi, y)$

        \STATE Compute logits $z^{(\text{aux})} = h_{\text{aux}}(\phi)$
        \STATE Compute Laplace covariance matrix $\Sigma$ using $h_{\text{aux}}$ and $\phi$
        \STATE Sample predictive distribution: $p = \mathbb{E}_{z \sim \mathcal{N}(z^{(\text{aux})}, \Sigma)} [\text{softmax}(z)]$
        \STATE Compute entropy margin: $\mathcal{H}(p) = -\sum_j p_j \log p_j$
        \STATE Compute weight $w = \exp(\beta \cdot \mathcal{H}(p)^\alpha)$

        \STATE Get teacher logits $z^{(T)} = T(x)$
        \STATE Get student logits $z^{(S)} = S(x)$
        \STATE Compute KD loss: $\mathcal{L}_{\text{KD}} = \text{KLDiv}(z^{(S)}, z^{(T)})$
        \STATE Compute CE loss: $\mathcal{L}_{\text{CE}} = \text{CE}(z^{(S)}, y)$
        \STATE Total loss: $\mathcal{L} = w \cdot \mathcal{L}_{\text{KD}} + \mathcal{L}_{\text{CE}}$
        \STATE Update student parameters via backprop on $\mathcal{L}$
    \ENDFOR
\ENDFOR
\STATE Save final student model and evaluate on test set
\end{algorithmic}
\end{algorithm}

The major differences between DeDIER and our approach, except for the implementation of Laplace Layer Approximaiton, can be listed as:
\begin{itemize}
    \item The auxiliary reweighting is done in every layer, since we are only training the student on five epochs and want to see the effect of the auxiliary network.
    \item The used auxiliary network is very simple, using only one layer. In our testing, last layer and third layer was used for separate test cases.

\end{itemize}

\subsubsection*{Models Used}
\begin{itemize}
    \item \textbf{Teacher model:} \texttt{bert-base-uncased} (12-layer BERT, hidden size 768)
    \item \textbf{Student model:} \texttt{distilbert-base-uncased} (6-layer DistilBERT, hidden size 768)
    \item \textbf{Auxiliary network:} One-layer linear classifier trained on the students third layer.
\end{itemize}

\subsubsection*{Optimization Hyperparameters}
\begin{itemize}
    \item \textbf{Teacher fine-tuning}
    \begin{itemize}
        \item Epochs: $E_T = 3$
        \item Learning rate: $2 \times 10^{-5}$
        \item Optimizer: AdamW
    \end{itemize}
    \item \textbf{Student training}
    \begin{itemize}
        \item Epochs: $E_S = 5$
        \item Learning rate: $2 \times 10^{-5}$
        \item Optimizer: AdamW
        \item Distillation temperature: $\tau = 2.0$
    \end{itemize}
\end{itemize}

\subsubsection*{Training and Evaluation Details}
\begin{itemize}
    \item Batch size: 16
    \item Dataset: MultiNLI (via HuggingFace \texttt{datasets})
    \item Evaluation: Accuracy and per-group performance (negation vs. non-negation)
    \item Tokenization: \texttt{bert-base-uncased} tokenizer with padding and truncation
    \item Learning rate scheduler: Linear schedule with warm-up
\end{itemize}

\subsubsection*{Evaluation Metrics}
The model is evaluated using accuracy over numerous groups, compared to the accuracy of the teacher model and the original DeDiER model. We also look at the confidence margins in the model in different layers of the algorithm to better understand how the change affects the model.

\section{Results}

\begin{figure}[H]
    \centering
    \begin{subfigure}[t]{0.45\linewidth}
        \centering
        \includegraphics[width=\linewidth]{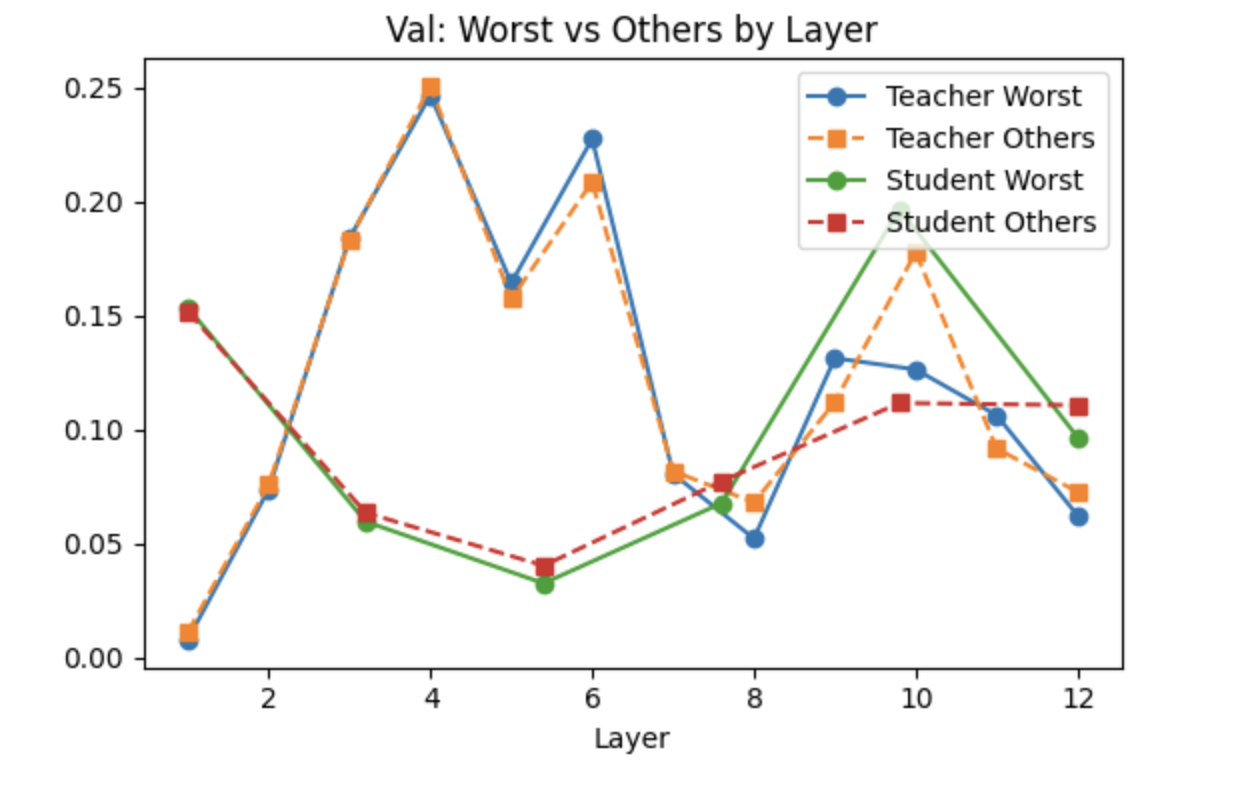}
        \caption{Auxiliary network on layer 3}
        \label{fig:aux3}
    \end{subfigure}
    \hfill
    \begin{subfigure}[t]{0.45\linewidth}
        \centering
        \includegraphics[width=\linewidth]{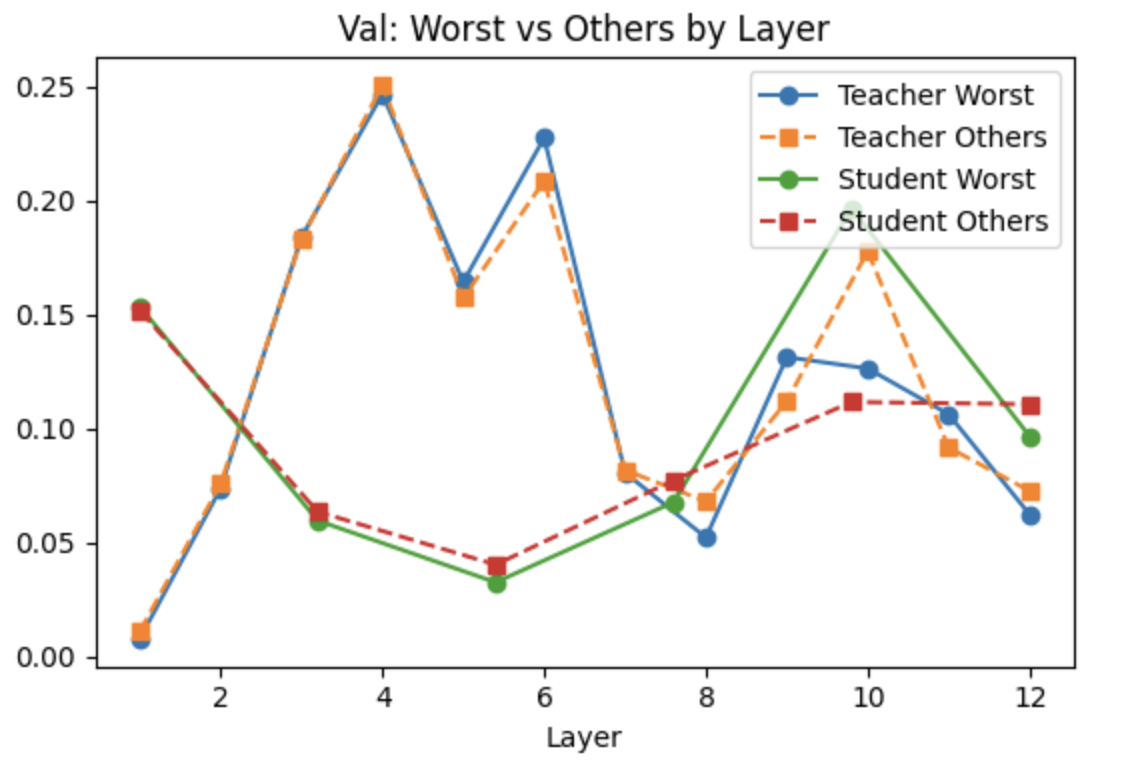}
        \caption{Auxiliary network on layer 6 (last layer)}
        \label{fig:aux6}
    \end{subfigure}
    \caption{Confidence margins per layer on worst group predictions and all predictions}
    \label{fig:layer-prediction}
\end{figure}

\begin{figure}[H]
    \centering
    \begin{subfigure}[t]{0.45\linewidth}
        \centering
        \includegraphics[width=\linewidth]{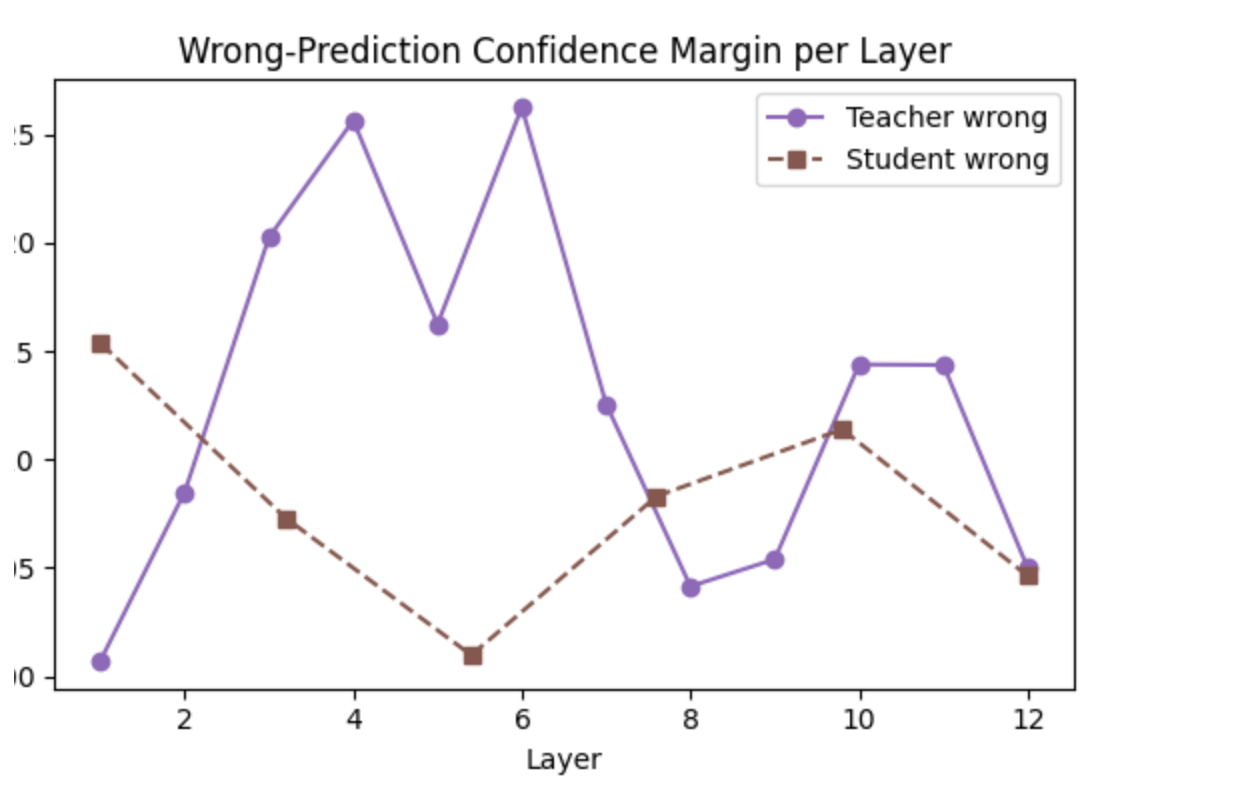}
        \caption{Auxiliary network on layer 3}
        \label{fig:aux3}
    \end{subfigure}
    \hfill
    \begin{subfigure}[t]{0.45\linewidth}
        \centering
        \includegraphics[width=\linewidth]{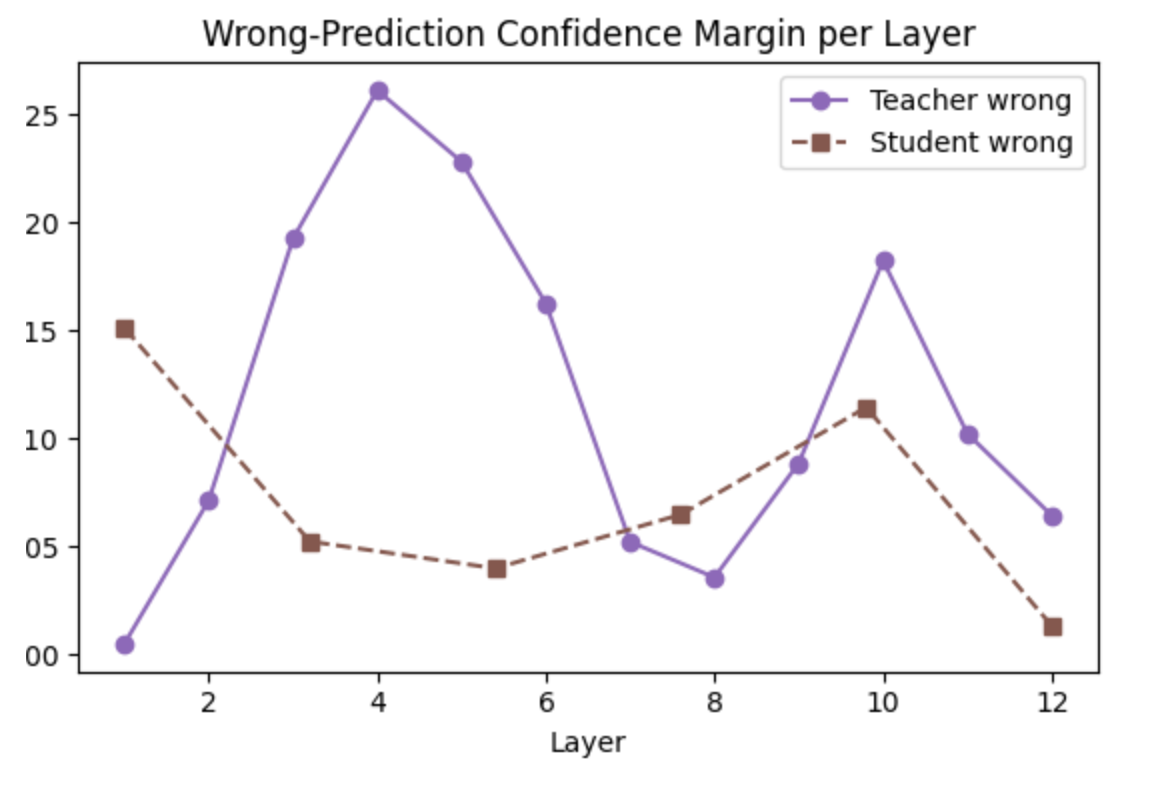}
        \caption{Auxiliary network on layer 6 (last layer)}
        \label{fig:aux6}
    \end{subfigure}
    \caption{Confidence margins in student and teacher on the predictions that were wrong}
    \label{fig:wrong-pred-con}
\end{figure}

The results are presented in Table \ref{fig:results table}, Table, \ref{fig:enter-label} Figure \ref{fig:layer-prediction} and Figure \ref{fig:wrong-pred-con}.

\subsection{Analysis}

\subsubsection{Improvement in Accuracy}
Looking at the results of the tests, a small improvement in accuracy of the student model is shown by the proposed approach compared to the original DeDIER model. This shows that the Laplace transformation has potential to enhance the results.

Using a single-layer auxiliary network on the third of six layers was found to be more effective than placing the auxiliary network on the final layer of the model. This observation aligns with the argument that earlier layers tend to capture more fundamental and informative representations. As a result, addressing the problem at an earlier stage may offer greater benefits by influencing the model’s internal feature processing more directly \cite{tiwari_using_2024}.

\subsubsection{Changes in confidence margin}

The confidence margins are generally lower than those reported in the original paper \cite{tiwari_using_2024}, as well as those of the teacher model when using the Laplace transformation. Even though only one layer was used for auxiliary, the effect of the reweighting affects multiple layers in both test cases. 

While lower confidence margins can sometimes indicate increased model uncertainty—which may be beneficial for calibration and avoiding overconfident incorrect predictions—they can also signal weaker separation between correct and incorrect classes, potentially reducing model reliability. Despite this ambiguity, the observed improvements in performance suggest that the approach holds promise. The results indicate potential for further gains with continued refinement of the auxiliary training and reweighting strategy.

\subsection{Future Work}

The Laplace transformation has shown potential to enhance the DeDIER model. However, to justify the results and enable further improvements, additional testing is required.

\subsubsection{More Advanced Auxiliary Layer}

The auxiliary network used in this study is a simple, single-layer model. To further evaluate the concept and assess the contribution of each layer in the network, a more sophisticated auxiliary architecture should be explored. Meronen et al. \cite{meronen2024fixing} propose combining predictions from auxiliary networks up to a given layer, as described in Eq. \ref{eq:culm}, where the weights $w_k$ represent the computational cost in FLOPs up to classifier $m$. While accounting for computational cost may be less important in the context of knowledge distillation, a similar approach for aggregating predictions across multiple layers could still be evaluated.

\begin{equation}
p_k^{\text{ens}}(\hat{\mathbf{y}}_i \mid \mathbf{x}_i) = \frac{1}{\sum_{l=1}^{k} w_l} \sum_{m=1}^{k} w_m \, p_m(\hat{\mathbf{y}}_i \mid \mathbf{x}_i)
\label{eq:culm}
\end{equation}

\subsubsection{More Datasets}

The current experiments are limited to the MultiNLI dataset. To gain a deeper understanding of the proposed model, it should be evaluated on additional datasets. Suitable datasets include Waterbirds, Civil Comments, and CelebA, which were used in the original paper and would allow for direct comparison. Moreover, the MultiNLI dataset itself offers further opportunities for analysis by exploring alternative subgrouping strategies when assessing worst-group performance.

\subsubsection{Hyperparameter Fine tuning}

Further work is needed to tune the hyperparameters of the proposed approach. This includes adjusting the number of training epochs, temperature parameters in the Laplace transformation, and the weighting scheme.

\begin{table}[h!]
\centering
\begin{tabular}{|l|c|c|c|c|}
\hline
Metric                          & Teacher & Student (Aux layer 3) & 
 Student (Aux layer 6) & Group \\
\hline
Average Accuracy                & 0.845   & 0.835 & 0.832  &  All   \\

\hline
\end{tabular}
\caption{\\ Results of Dedier with Laplace}
\label{fig:results table}
\end{table}

\begin{table}[h!]
\centering
\begin{tabular}{|l|c|c|}
\hline
Metric             & Teacher & Student \\
\hline
Final Test Accuracy & 0.841   & 0.830   \\
\hline
\end{tabular}
\caption{\\ Original DEDIER performance with same teacher}
\label{fig:enter-label}
\end{table}



\section{Conclusion \label{sec: conc}}
We conclude that Laplace-based approximation for early exit layers is a cheap yet effective way to compute uncertainty estimates from early layers, which can be used to reweight the KD loss in student models and make their predictions less reliant on simple features.



\bibliographystyle{plainnat}
\bibliography{KD_early_exit}

\appendix

\section{Appendix / supplemental material}

\subsection{Code}

The code used can be found on \url{https://github.com/addvinf/DEDIER-with-Laplace/tree/main}

\end{document}